\documentclass{article} 
\usepackage{iclr2026_conference,times}
\iclrfinalcopy
\usepackage{graphicx} 
\usepackage{multirow} 
\usepackage{booktabs}
\usepackage{wrapfig}

\usepackage{amsmath,amsfonts,bm}

\def\eqref#1{equation~\ref{#1}}

\def\1{\bm{1}}

\DeclareMathAlphabet{\mathsfit}{\encodingdefault}{\sfdefault}{m}{sl}
\SetMathAlphabet{\mathsfit}{bold}{\encodingdefault}{\sfdefault}{bx}{n}

\usepackage{hyperref}
\usepackage{url}
\usepackage{subcaption}
\usepackage[most]{tcolorbox} 

\title{MAGE: Meta-Reinforcement Learning for Language Agents toward Strategic Exploration and Exploitation}

\author{
  Lu Yang\thanks{Equal contribution: \texttt{yanglu25@mails.tsinghua.edu.cn, zelai.eecs@gmail.com}} , 
  Zelai Xu\footnotemark[1] , 
  Minyang Xie, 
  Jiaxuan Gao, 
  Zhao Shok, 
  Yu Wang\thanks{Corresponding authors: \texttt{yu-wang@tsinghua.edu.cn, jxwuyi@gmail.com}} , 
  Yi Wu\footnotemark[2] \\
  Tsinghua University \\
}

\begin{document}

\maketitle
\ificlrfinal
    \thispagestyle{fancy}
    \lhead{Preprint. Under Review.}
\fi

\begin{abstract}
Large Language Model (LLM) agents have demonstrated remarkable proficiency in learned tasks, yet they often struggle to adapt to non-stationary environments with feedback. While In-Context Learning and external memory offer some flexibility, they fail to internalize the adaptive ability required for long-term improvement. Meta-Reinforcement Learning (meta-RL) provides an alternative by embedding the learning process directly within the model. However, existing meta-RL approaches for LLMs focus primarily on exploration in single-agent settings, neglecting the strategic exploitation necessary for multi-agent environments. We propose MAGE, a meta-RL framework that empowers LLM agents for strategic exploration and exploitation. MAGE utilizes a multi-episode training regime where interaction histories and reflections are integrated into the context window. By using the final episode reward as the objective, MAGE incentivizes the agent to refine its strategy based on past experiences. We further combine population-based training with an agent-specific advantage normalization technique to enrich agent diversity and ensure stable learning. Experiment results show that MAGE outperforms existing baselines in both exploration and exploitation tasks. Furthermore, MAGE exhibits strong generalization to unseen opponents, suggesting it has internalized the ability for strategic exploration and exploitation. 
Code is available at \url{https://github.com/Lu-Yang666/MAGE}.

\end{abstract}

\section{Introduction}
\label{sec:intro}

Reinforcement Learning (RL) has significantly enhanced the ability of Large Language Model (LLM) agents to solve complex, multi-step tasks~\cite{deepseekr1, gemini2.5}. 
However, despite their proficiency in static environments, these agents typically lack the capacity to incorporate real-time feedback or refine their strategies when encountering shifting dynamics. 
Bridging this gap requires a transition from fixed execution to test-time adaptation.
By enabling agents to ``learn from experience'' during interaction, their behavior can evolve dynamically across several trials. 
This capability is essential for transforming static task-solvers into adaptive learners that can autonomously adjust to the complexities of non-stationary environments.

\begin{figure}[!h]
    \centering
    \includegraphics[width=0.6\linewidth]{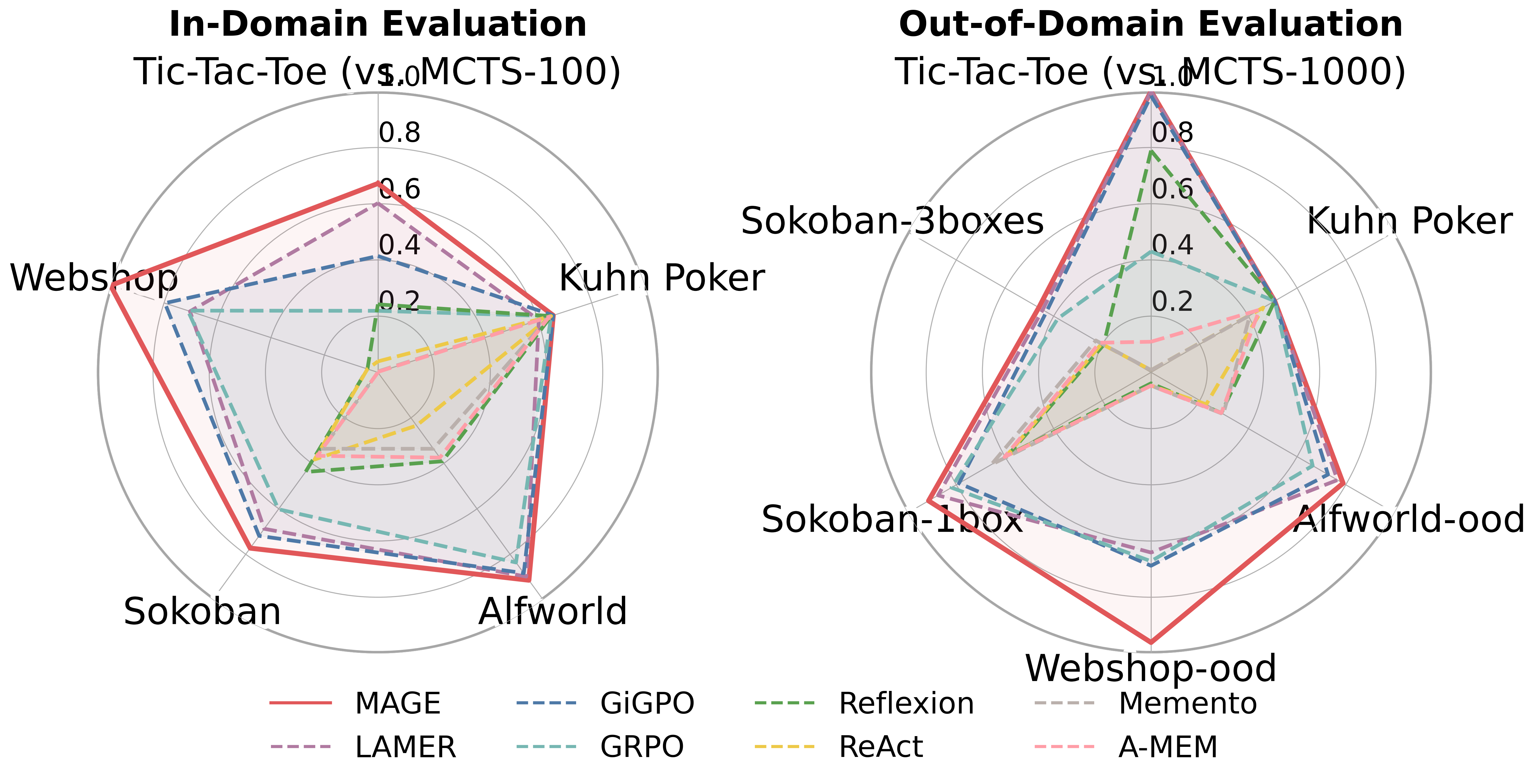}\caption{Evaluation of MAGE in
exploration and exploitation tasks. MAGE outperforms existing training and training-free baselines in a number of environments.
}
    \label{fig:teaser}
\end{figure}
Current approaches to achieve such test-time adaptation primarily rely on prompting through In-Context Learning (ICL)~\cite{gpt3, reflexion} or external memory components~\cite{xu2025mem, memento}.
Although these methods provide some flexibility, they often fail to fundamentally internalize the adaptive capability. Meta-reinforcement learning (meta-RL)~\cite{rl2, maml} offers a more robust alternative by embedding the learning mechanism directly within the model itself.
While recent efforts have applied meta-RL to LLM agents~\cite{lamer}, they focus almost exclusively on inducing exploration in single-agent tasks. 
In multi-agent environments, however, an agent must adapt not only to the environment but also to the diverse behaviors of other participants. 
This necessitates strategic exploitation, which involves dynamically identifying and capitalizing on opponent-specific patterns. 
Since a strategy optimized for one adversary might fail against another~\cite{czarnecki2020real}, this requires a fundamental shift from environment exploration to agent exploitation.

To bridge this gap, we present \textbf{MAGE}, a \textbf{\underline{M}}eta-RL framework that empowers LLM \textbf{\underline{A}}gents for strate\textbf{\underline{G}}ic Exploration and  \textbf{\underline{E}}xploitation in multi-agent environments.
MAGE adopts a multi-episode training regime where the agent's interaction history and reflections from previous episode are integrated into the LLM’s context window.
By optimizing the policy across these interaction trajectories via RL, the LLM internalizes the ability to learn from past experience.
Unlike prior work~\cite{lamer} that focuses on maximizing cumulative rewards to incentivize exploration, MAGE prioritizes the final episode reward as the primary objective. This metric serves as the measure of how effectively an agent has identified the weaknesses of its adversary and adapted its strategy to maximize performance. The core idea of MAGE is to transform the LLM agent into a strategic learner that views its interaction history not just as a record of events, but as the strategic basis for exploiting the opponents' vulnerabilities.

Meta-RL in multi-agent environments for strategic explotation and exploitation presents unique challenges, as training against a single agent is insufficient for developing generalizable exploitation skills and fails to prepare the model for the varied flaws found in diverse opponents~\cite{czarnecki2020real}. 
Therefore, we leverage population-based training (PBT)~\cite{jaderberg2017population, alphastar} to expose the LLM to a diverse pool of different agents. 
By interacting with a population of agents, MAGE learns to recognize specific behavioral patterns and exploit their inherent strategic vulnerabilities. 
We further introduce an agent-specific advantage normalization technique to handle divergent reward distributions across opponents, which necessitate separate normalization. 
Our agent-specific normalization ensures that the agent can effectively use its context window as a strategic buffer to distinguish between different types of adversaries and respond with the appropriate counter-strategy. 
By integrating population-based training with agent-specific normalization, MAGE establishes a principled meta-RL framework for learning strategic exploration and exploitation in language agents.

We conduct extensive experiments to validate MAGE across diverse benchmarks. In single-agent (Alfworld, Webshop, Sokoban) and multi-agent (Tic-Tac-Toe, Kuhn Poker) settings, MAGE consistently outperforms baselines with a rapid adaptation curve. Notably, it achieves a $100\%$ success rate in Webshop and $91.4\%$ in Alfworld, significantly surpassing the strongest baselines of $79.7\%$ and $88.3\%$. In Tic-Tac-Toe, MAGE reaches a $67.2\%$ success rate, exceeding LAMER's $60.2\%$. Evaluation against unseen opponents further demonstrates robustness: MAGE achieves $96.1\%$ in Webshop-ood (vs. $68.8\%$), maintains top performance in Sokoban, and reaches a $100.0\%$ draw rate against MCTS-1000 in Tic-Tac-Toe. In Kuhn Poker, it hits the theoretical upper bound against CFR opponents. These results suggest MAGE internalizes a fundamental logic for zero-shot adaptation rather than mere pattern memorization. Finally, ablation studies confirm that the synergy between population-based training and agent-specific normalization is crucial for identifying and exploiting opponent vulnerabilities.
We summarize our contributions as follows:
\begin{itemize}
    \item We propose MAGE, a meta-RL framework that empowers language agents for strategic exploration and exploitation in multi-agent environments.
    \item We introduce an effective training recipe that combines population-based training with agent-specific advantage normalization to provide diverse opponents and stable training signals for meta-RL.
    \item We conduct extensive experiments showing that MAGE improves adaptation and achieves higher win rates against in-domain and unseen opponents.
\end{itemize}


\section{Related Work}
\label{sec:related}
\paragraph{In-Context Learning}
The capacity of Large Language Models (LLMs) to adapt via In-Context Learning (ICL)~\cite{gpt3} has catalyzed the development of autonomous agents. To move beyond static prompting, recent works such as \textit{Reflexion}~\cite{reflexion} and \textit{Self-Refine}~\cite{selfrefine} introduce iterative feedback loops, allowing agents to correct errors based on environment trials. Furthermore, memory-augmented frameworks~\cite{xu2025mem, memento} enable agents to retrieve past experiences from external databases. However, these methods rely on fixed model weights and often fail to internalize the underlying learning logic, leading to sub-optimal adaptation in complex, non-stationary settings.

\paragraph{Agentic RL for LLMs}
Reinforcement Learning (RL) has shifted from simple preference alignment to enhancing complex reasoning and multi-turn decision-making. High-profile models like OpenAI o1~\cite{o1} and DeepSeek-R1~\cite{deepseekr1} demonstrate the power of RL in scaling computational thought chains. In the agentic context, RL is increasingly applied to multi-step tasks such as web search~\cite{searchr1}, software engineering~\cite{swerl}, and GUI interactions~\cite{wang2025ui}. Specialized algorithms like GiGPO~\cite{gigpo} have been proposed to stabilize training for such long-horizon interaction trajectories. MAGE builds upon this agentic RL trend but shifts the focus from mastering a single task to mastering the process of adaptation itself.

\paragraph{Meta-Reinforcement Learning}
Traditional Meta-RL~\cite{rl2, maml} aims to train agents that can rapidly adapt to new tasks by internalizing the learning procedure. Recently, this paradigm has been extended to LLMs; for instance, LAMER~\cite{lamer} utilizes meta-RL to incentivize efficient exploration in single-agent environments. However, in multi-agent or competitive scenarios, agents must perform \textit{strategic exploitation}—identifying and capitalizing on the specific vulnerabilities of opponents. Unlike previous works that focus on exploration, MAGE leverages population-based training and agent-specific normalization to ensure the agent can robustly exploit diverse behaviors, representing a novel frontier for meta-RL in language agents.

\section{MAGE}
\label{sec:method}
\begin{figure*}[t]
\centering
\includegraphics[width=0.85\linewidth]{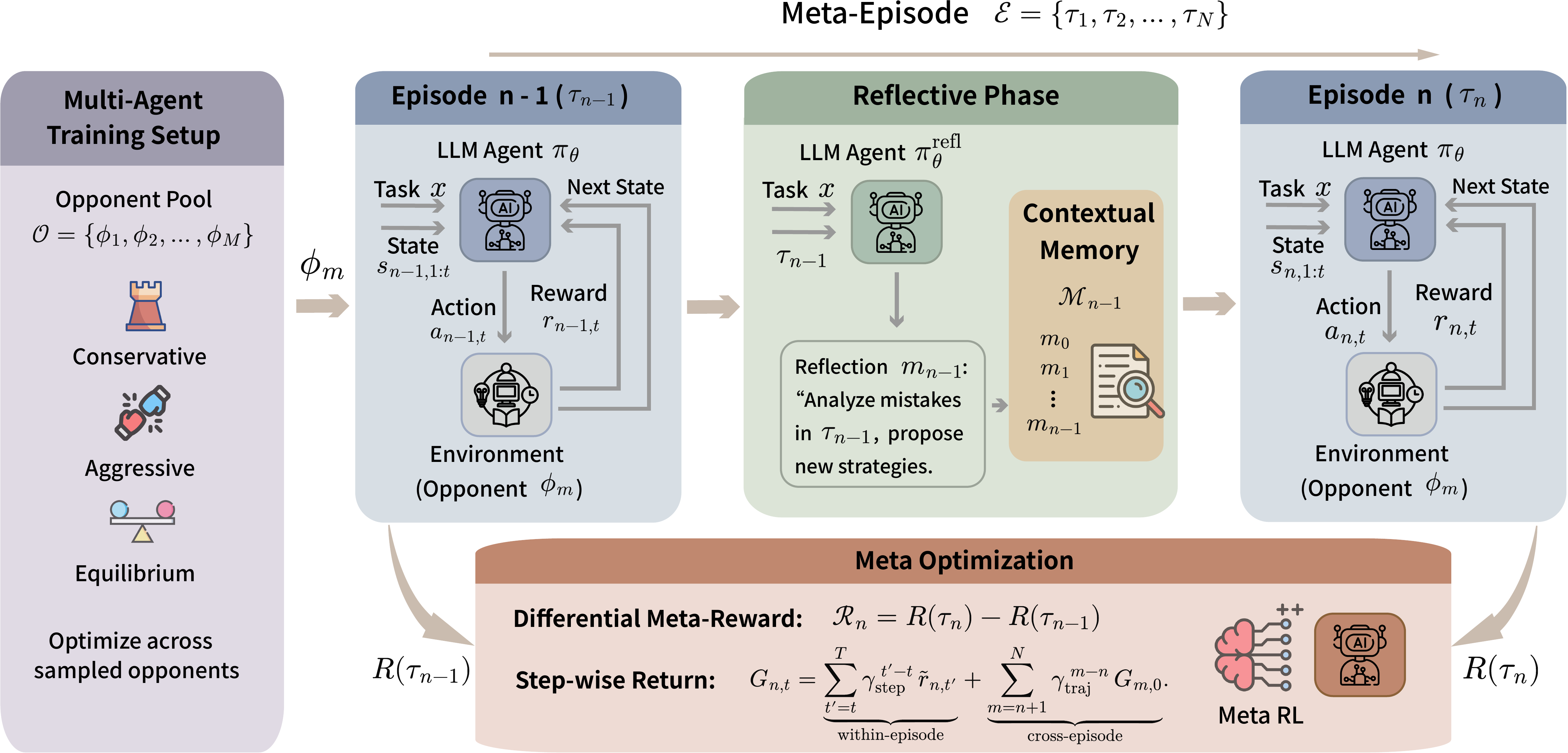}
\caption{Overview of the MAGE framework. 
MAGE optimizes an LLM policy $\pi_\theta$ across $N$ episodes using a contextual memory $\mathcal{M}_{n-1}$ updated via self-reflections $m_{n-1}$. By training against an opponent pool $\mathcal{O}$ and applying agent-specific advantage normalization to the final episode reward, MAGE facilitates stable strategic adaptation and learning-to-learn.
}
\label{fig:overview}
\end{figure*}

In this section, we present \textbf{MAGE}, a \textbf{\underline{M}}eta-RL framework designed to optimize LLM \textbf{\underline{A}}gents for strate\textbf{\underline{G}}ic Exploration and \textbf{\underline{E}}xploitation in multi-agent environments. Unlike standard In-Context Learning (ICL), which relies on emergent behaviors, MAGE explicitly trains the model to \emph{learn to learn} by treating a sequence of interaction episodes as an inner optimization loop.
\subsection{Problem Setup}
We define a \emph{meta-episode} as a sequence of $N$ episodes
\[
\mathcal{E} = \{\tau_1, \tau_2, \dots, \tau_N\},
\]
executed against a stationary task or opponent. Each episode $\tau_n$ corresponds to a complete trajectory:
\begin{equation}
\tau_n =
\{(s_{n,1}, a_{n,1}, r_{n,1}), \dots, (s_{n,T}, a_{n,T}, r_{n,T})\},
\end{equation}
where $s_{n,t}$, $a_{n,t}$, and $r_{n,t}$ denote the state, action, and reward at step $t$ of episode $n$, respectively.

\subsection{The MAGE Framework}
Unlike standard ICL, our framework introduces a \emph{Reflective Inner Loop} in which the model explicitly generates and exploits its own high-level feedback across episodes. At the conclusion of episode $\tau_{n-1}$, the model produces a self-generated reflection:
\begin{equation}
m_{n-1} \sim \pi_\theta^{\text{refl}}(\cdot \mid \tau_{n-1}, x),
\end{equation}
where $\pi_\theta^{\text{refl}}$ denotes the model’s reflection-generation behavior and $x$ denotes the specific task description.
The reflection $m_{n-1}$ is expressed in natural language and is intended to summarize failure modes, diagnose strategic errors, and propose corrective actions. 

The sequence of past reflections is organized into a \emph{contextual memory}:
\begin{equation}
\mathcal{M}_{n-1} = \{m_0, m_1, \dots, m_{n-1}\},
\end{equation}
which serves as a compact, high-level abstraction of accumulated experience across episodes. 

Within episode $n$, the model generates actions based on the task description, the contextual memory formed by all prior reflections, and the state history observed so far in the current episode. Formally, let
\[
s_{n,1:t} = \{s_{n,1}, \dots, s_{n,t}\}
\]
denote the state history up to step $t$ in episode $n$. The action distribution is then defined as:
\begin{equation}
a_{n,t} \sim \pi_\theta\!\left(\cdot \mid s_{n,1:t}, \mathcal{M}_{n-1}, x \right).
\end{equation}

For the initial episode ($n=1$), $\mathcal{M}_0 = \{m_0\}$ is initialized as an empty reflection.

\subsubsection{Step-wise Return}
To facilitate fine-grained policy updates, we compute a step-wise return \(G_{n,t}\) for each action \(a_{n,t}\). In our setting, the environment provides a \emph{sparse} reward: all task-related reward is emitted only at the final step of each episode. Consequently, the learning signal is defined primarily at the episode level, and is injected into the step-wise return in a structured manner.

\paragraph{Episode-wise differential meta-reward}

To explicitly optimize learning progress across episodes, we define the episode-wise differential meta-reward:
\begin{equation}
\mathcal{R}_n = R(\tau_n) - R(\tau_{n-1}),
\end{equation}
with \(R(\tau_0)\equiv 0\). \(R(\tau_n)\) denotes the cumulative task reward of episode \(n\). This signal measures the improvement achieved by the policy update induced by the reflection from the previous episodes.

The step-wise reward is then formulated as:
\begin{equation}
\tilde{r}_{n,t} =
\begin{cases}
0, & t < T, \\[6pt]
\mathcal{R}_n, & t = T.
\end{cases}
\end{equation}

Inspired by LaMER, the step-wise return for action \(a_{n,t}\) is defined as:
\begin{equation}
G_{n,t}
=
\underbrace{
\sum_{t' = t}^{T}
\gamma_{\mathrm{step}}^{\,t' - t}
\, \tilde{r}_{n,t'}
}_{\text{within-episode}}
+
\underbrace{
\sum_{m = n+1}^{N}
\gamma_{\mathrm{traj}}^{\,m - n}
\, G_{m,0}
}_{\text{cross-episode}} .
\end{equation}
where $\gamma_{\mathrm{step}}$ and $\gamma_{\mathrm{traj}}$ denote the discount factors
for within-episode and cross-episode returns, respectively. For more information on the composition of episode reward, see Appendix~\ref{app:reward-design}.

\subsubsection{Optimization Objective}

The overall objective of MAGE is to maximize the expected cumulative meta-reward across a meta-episode:
\begin{equation}
\max_{\theta} \;
\mathbb{E}_{\tau_1, \dots, \tau_N \sim \pi_{\theta}}
\left[
\sum_{n=1}^{N} \mathcal{R}_n
\right],
\end{equation}
where $\pi_{\theta}$ denotes the LLM policy parameterized by $\theta$.

Using the derived advantages $\hat{A}_{n,t}$ calculated from the step returns $G_{n,t}$, the policy is optimized via a generalized policy gradient objective. By utilizing the advantage as a weighting factor for the log-probabilities of actions conditioned on both the state history and the contextual memory, the meta-learning objective is formulated as:
\begin{equation}
\mathcal{L}_{\text{episode}}(\theta)
=
- \sum_{n=1}^{N} \sum_{t=1}^{T}
\hat{A}_{n,t}
\log \pi_{\theta}(a_{n,t} \mid s_{n,1:t}, \mathcal{M}_{n-1}, x),
\end{equation}

\begin{equation}
\mathcal{L}_{\text{MAGE}}(\theta)
=
\mathbb{E}_{\mathcal{E} \sim \pi_\theta} 
\left[
\mathcal{L}_{\text{episode}}(\theta)
\right].
\end{equation}

This objective explicitly encourages the model to acquire strategies that improve its own learning dynamics over successive episodes.
It is worth noting that the MAGE framework is algorithm-agnostic; while the basic objective is presented as a policy gradient, it is fully compatible with advanced reinforcement learning algorithms such as GiGPO ~\cite{gigpo}. In these cases, the advantage $\hat{A}_{n,t}$ serves as the core signal for the respective surrogate loss functions, effectively training the model to optimize its strategy refinement process across the meta-episode.

\subsubsection{Agent-specific Advantage Normalization}
In multi-agent setting, the LLM agent must identify and exploit the behavioral patterns of diverse opponents.
We implement two games (\textit{Tic-Tac-Toe}, \textit{Kuhn Poker}) that represent different strategic challenges.

During the training phase, the LLM agent interacts with a population of opponents $\mathcal{O} = \{\phi_1, \phi_2, \dots, \phi_M\}$, where each $\phi_m$ represents a distinct fixed strategy. A meta-episode consists of $N$ episodes against a single opponent $\phi_m$ sampled from $\mathcal{O}$. The agent is not explicitly told which strategy it is facing; instead, it must infer the opponent's play pattern from the contextual memory $\mathcal{M}_{n-1}$ 
and the current episode's state history $s_{n,1:t}$, and adjust its policy $\pi_\theta$ accordingly.

Let $\hat{A}_{n,t}^{(m)}$ denote the normalized step-wise advantage derived
from interactions with opponent $\phi_m$. The multi-agent MAGE objective
maximizes the expected advantage-weighted log-likelihood of actions
across all steps and episodes within the meta-episode, and across sampled opponents:
\begin{equation}
\mathcal{L}_{\phi_m}(\theta)
=
- \sum_{n=1}^{N} \sum_{t=1}^{T}
\hat{A}_{n,t}^{(m)}
\log \pi_\theta
\bigl(a_{n,t} \mid s_{n,1:t}, \mathcal{M}_{n-1}, x \bigr),
\end{equation}

\begin{equation}
\mathcal{L}_{\mathrm{MAGE}}^{\mathrm{multi-agent}}(\theta)
=
\mathbb{E}_{\phi_m \sim \mathcal{O}}
\left[
\mathcal{L}_{\phi_m}(\theta)
\right].
\end{equation}

This objective encourages the agent to select actions that maximize learning progress
against a heterogeneous population of opponents, effectively promoting
strategy adaptation and generalization across different play patterns.
\begin{table*}[t]
\centering
\small
\setlength{\tabcolsep}{3pt}

\begin{tabular}{llccccc}
\toprule
\multirow{2}{*}{\textbf{Category}} 
& \multirow{2}{*}{\textbf{Method}}
& \multicolumn{2}{c}{\textbf{Multi-Agent Tasks}}
& \multicolumn{3}{c}{\textbf{Single-Agent Tasks}} \\
\cmidrule(lr){3-4} \cmidrule(lr){5-7}

& 
& \textbf{Kuhn Poker}
& \textbf{Tic-Tac-Toe}
& \textbf{ALFWorld}
& \textbf{Sokoban}
& \textbf{WebShop} \\
\midrule

\multirow{2}{*}{In-Context Learning}
& ReAct       & 0.648 & 0.039 & 0.234 & 0.383 & 0.039 \\
& Reflexion   & 0.648 & 0.242 & 0.391 & 0.438 & 0.039 \\

\midrule
\multirow{2}{*}{Memory-based}
& A-MEM       & 0.641 & 0.016 & 0.375 & 0.367 & 0.000 \\
& Memento     & 0.641 & 0.031 & 0.336 & 0.336 & 0.000 \\

\midrule
\multirow{2}{*}{RL}
& GRPO        & 0.648 & 0.219 & 0.836 & 0.602 & 0.711 \\
& GiGPO       & \textbf{0.656} & 0.414 & 0.883 & 0.719 & 0.797 \\

\midrule
\multirow{2}{*}{Meta-RL}
& LAMER       & 0.594 & 0.602 & 0.898 & 0.688 & 0.703 \\
& \textbf{MAGE}
               & \textbf{0.656} & \textbf{0.672}
               & \textbf{0.914} & \textbf{0.773} & \textbf{1.000} \\

\bottomrule
\end{tabular}

\caption{
\textbf{In-domain evaluation performance of the final episode.}
MAGE consistently reaches or outperforms existing methods.
}
\label{tab:indomain_full}
\end{table*}

\section{Experiments}
\label{sec:exp}




\subsection{Experimental Setup}
We employ Qwen3-4B as our base large language model, utilizing its native Thinking capabilities to facilitate complex reasoning during both the reflection and action generation phases.The training is conducted using the GiGPO algorithm. For the meta-learning objective, we set the cross-episode discount factor $\gamma_{\mathrm{traj}} = 0.6$. Each meta-episode consists of $N=3$ episodes. During MAGE training, we utilize a group size of 8 meta-episodes per batch. In contrast, for the standard RL baselines, we expand the group size to 24 to maintain an equivalent number of total trajectories per update, ensuring a fair comparison of sample efficiency.
In multi-agent environments, we utilize population-based training: for Tic-Tac-Toe, the agent interacts with MCTS-based, preferred-pattern, and random strategies; for Kuhn Poker, the training distribution consists of conservative, aggressive, and intermediate opponent archetypes. 

\paragraph{Environments.} We evaluate MAGE across diverse strategic benchmarks. Multi-agent tasks include \textit{Tic-Tac-Toe}, a perfect-information game for assessing rapid adaptation to deterministic optimal play, and \textit{Kuhn Poker}, an imperfect-information variant testing strategic reasoning and bluffing. Single-agent tasks include \textit{ALFWorld} (interactive household planning), \textit{WebShop} (goal-oriented web navigation), and \textit{Sokoban} (long-horizon spatial puzzles). 

\paragraph{Baselines.}
We compare MAGE against a broad set of baselines encompassing heuristic agent frameworks (\textit{ReAct}~\cite{react}, \textit{Reflexion}~\cite{reflexion}), memory-augmented agents (\textit{A-MEM}~\cite{xu2025mem}, \textit{Memento}~\cite{memento}), foundational reinforcement learning methods (\textit{GRPO}~\cite{grpo}, \textit{GiGPO}~\cite{gigpo}), and related meta-learning approaches (\textit{LAMER}~\cite{lamer}). 

\paragraph{Metrics.} The primary evaluation metric across all environments is the \emph{Success Rate}, reported under the \emph{Pass@k} formulation. Specifically, Pass@k measures the probability that an agent successfully completes the task at least once within the first $k$ episodes of a meta-episode. 





Our evaluation consists of three parts to systematically assess strategic plasticity and the Final-Episode Optimization objective:

\begin{itemize} \item \textbf{In-Domain Evaluation:} Assesses fundamental learn-to-learn performance and the conversion of early interactions into exploitative strategies under training-consistent distributions. \item \textbf{Generalization and Cross-Domain Plasticity:} Tests robustness against out-of-domain (OOD) tasks and unseen opponents to verify a generalizable probing logic rather than pattern memorization. \item \textbf{Ablation Studies:} Deconstructs the framework to analyze the impact of Final-Episode Optimization, Population-Based Training (PBT), and Opponent-Specific Advantage Normalization.
\end{itemize}

\subsection{In-Domain Evaluation}
\begin{wrapfigure}[33]{r}{0.5\textwidth}
  \centering
  
  \begin{minipage}{\linewidth}
    \centering
    \includegraphics[width=0.95\linewidth]{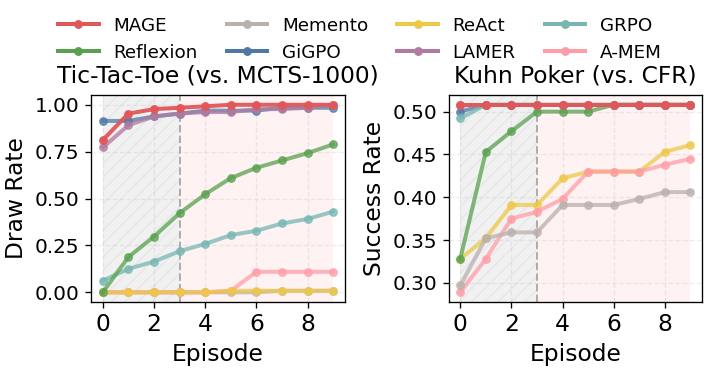}
    \caption{\textbf{Multi-Agent Evaluation.} Performance in Tic-Tac-Toe (vs. MCTS-1000) and Kuhn Poker (vs. CFR).}
    \label{fig:multi_agent}
  \end{minipage}

  \vspace{1.5em} 

  \begin{minipage}{\linewidth}
    \centering
    \includegraphics[width=0.95\linewidth]{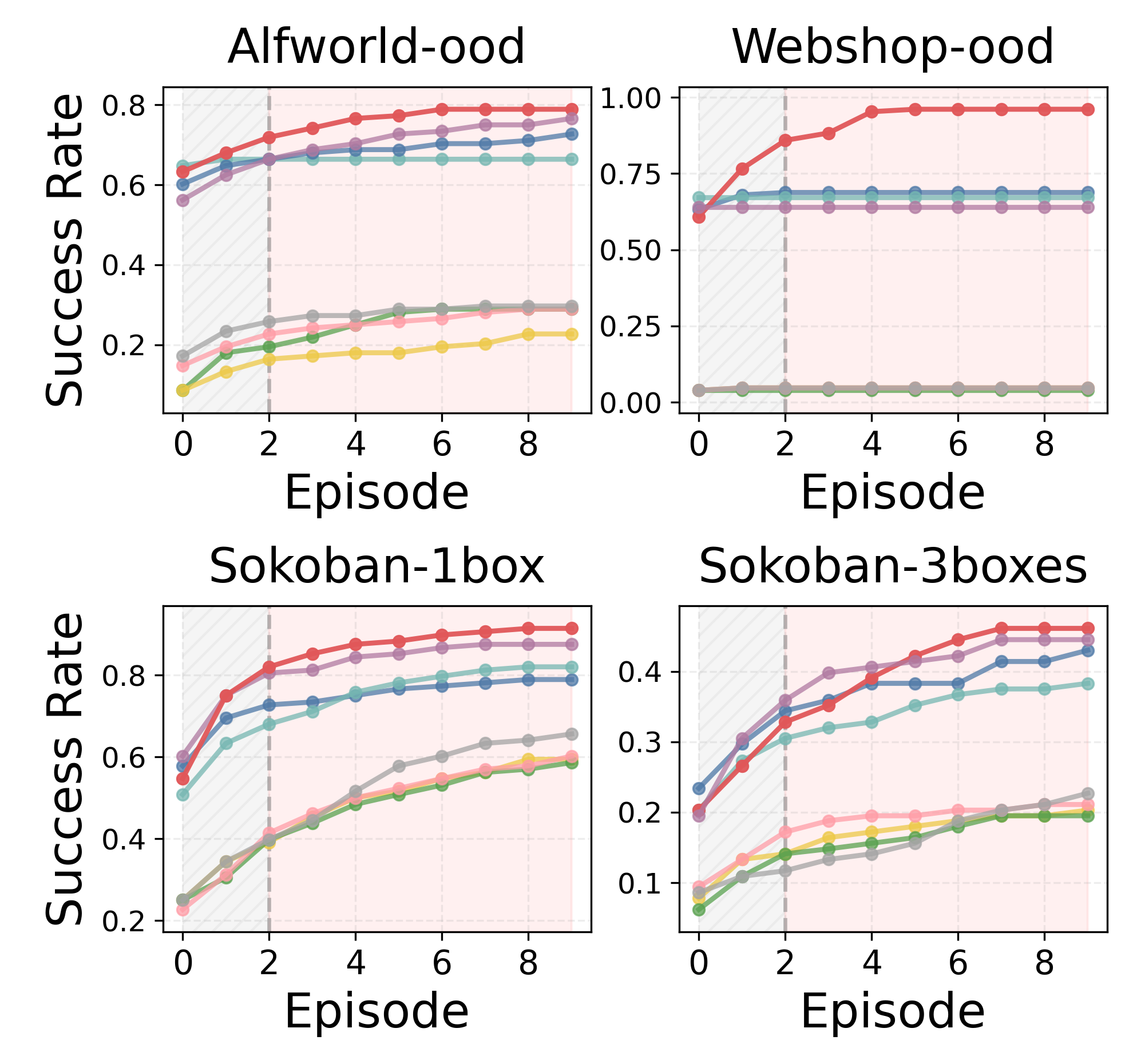}
    \caption{\textbf{Single-Agent Evaluation.} Performance in ALFWorld, Sokoban, and WebShop.}
    \label{fig:oodsingle_agent}
  \end{minipage}
  
\end{wrapfigure}
We evaluate the effectiveness of MAGE under in-domain conditions, where evaluation environments and opponent distributions align with the training phase. This isolates the impact of the proposed Final-Episode Optimization objective and its ability to foster strategic plasticity in LLM agents. The results are shown in Table~\ref{tab:indomain_full}.

Our training objective explicitly optimizes the final episode return within a three-episode trajectory. Consequently, the policy is not incentivized to maximize rewards in the initial two episodes, where performance may naturally trail baselines. The effectiveness of MAGE is best observed from the third episode onward, as this phase reflects the completed adaptation process targeted by our objective. Therefore, we emphasize performance in the third episode and beyond to provide a faithful evaluation of the method's strategic optimization.
\paragraph{Multi-Agent Strategic Exploitation.}
The evaluation in multi-agent settings (Tic-Tac-Toe and Kuhn Poker) highlights MAGE's capacity for \textbf{strategic exploitation} of opponent-specific idiosyncrasies.

In Tic-Tac-Toe, MAGE achieves a dominant $67.2\%$ terminal success rate against MCTS-100, significantly outperforming LAMER ($60.2\%$) and GiGPO ($41.4\%$). 
In Kuhn Poker, MAGE hits the $65.6\%$ theoretical upper bound, matching the performance ceiling despite the task's stochasticity. These results validate that Opponent-Specific Advantage Normalization effectively stabilizes meta-learning across heterogeneous strategy populations.


\paragraph{Single-Agent Exploration.}

In complex single-agent tasks (ALFWorld, Sokoban, WebShop), MAGE consistently achieves superior terminal performance, proving that prioritizing terminal success over cumulative reward fosters more effective \emph{in-context adaptation}.

In WebShop, MAGE transitions from a $66.4\%$ initial success rate to \textbf{$100\%$ by the 5th episode}, outperforming baselines like GiGPO and LAMER by $20-30\%$. This highlights its ability to convert early feedback into flawless execution. In Sokoban, MAGE demonstrates a ``slow-start, high-finish" pattern, improving from $40.6\%$ to \textbf{$77.3\%$} ($+36.7\%$). This confirms that Final-Episode Optimization incentivizes strategic probing over conservative play. In ALFWorld, MAGE reaches a \textbf{$91.4\%$} Pass@10, surpassing LAMER ($89.8\%$) and GiGPO ($88.3\%$), while leaving pure prompting methods like Reflexion below $40\%$.\paragraph{Discussion on Meta-Learning Capability.}Unlike static methods (ReAct, Reflexion) that fail to improve across episodes, MAGE treats interaction history as a ``meta-context." By optimizing for terminal success, it systematically transitions from early \emph{information-gathering} to late-episode \emph{exploitation}, achieving true strategic plasticity.

\subsection{Generalization and Cross-Domain Plasticity}
We evaluate MAGE under out-of-domain (OOD) conditions to assess its strategic exploitation across shifted task complexities and novel opponent behaviors.\paragraph{Multi-agent Evaluation.} Facing MCTS-1000 in Tic-Tac-Toe—where winning is nearly impossible—MAGE's draw rate ascends from $81.2\%$ to \textbf{$100.0\%$} by the final episode. This demonstrates its ability to identify perfect defensive patterns and recalibrate to prevent exploitation. Against CFR opponents in Kuhn Poker, MAGE reaches the \textbf{$50.8\%$} theoretical ceiling. This stability validates that Opponent-Specific Advantage Normalization prevents policy collapse when encountering near-optimal behaviors.\paragraph{Single-agent Evaluation.} In Sokoban, despite being trained only on 2-box configurations, MAGE achieves \textbf{$91.4\%$} in 1-box and \textbf{$46.1\%$} in 3-boxes variants, outperforming GiGPO. In WebShop, MAGE maintains a \textbf{$96.1\%$} success rate (vs. $68.8\%$ for GiGPO), and in AlfWorld, it preserves a high terminal performance of \textbf{$78.9\%$}. These results suggest MAGE develops a robust information-gathering mechanism that generalizes to idiosyncratic features under distributional shifts rather than merely memorizing patterns.

\subsection{Ablation Studies}
We conduct controlled ablations on MAGE's core components: reward design, population-based training, and agent-specific advantage normalization, maintaining fixed hyperparameters and budgets.
\subsubsection{Reward Design}
\begin{wrapfigure}[17]{r}{0.45\textwidth} 
  \centering
  \includegraphics[width=0.42\textwidth]{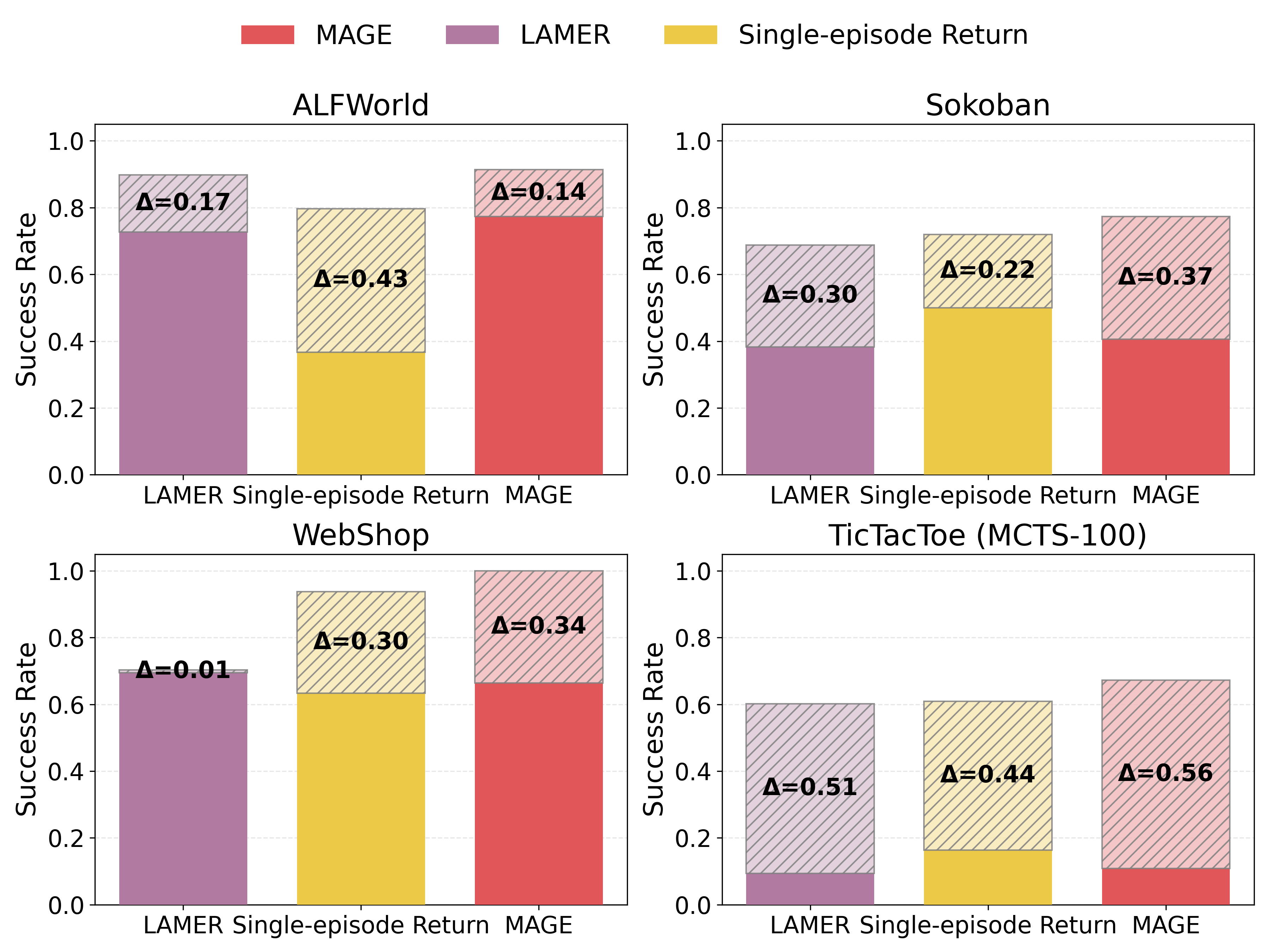}
  \caption{\textbf{Reward design ablation results.} Using single-episode return and LAMER variants causes the success rate to drop in various environments.}
  \label{fig:ablation_reward_design}
\end{wrapfigure}
We compare three reward formulations while preserving the meta-episode structure:
\textbf{Differential Return (MAGE):} Uses episode-wise progress, $\hat r_{n,T}= R(\tau_n) - R(\tau_{n-1})$. \textbf{Cumulative Return (LAMER-style):} Uses absolute performance, $\hat r_{n,T} = R(\tau_n)$, with cross-episode propagation. \textbf{Single-episode Return:} Uses $\hat r_{n,T} = R(\tau_n)$ without cross-episode propagation.

Results in Figure~\ref{fig:ablation_reward_design} show that MAGE's \textbf{Differential Return} is the primary driver for steep learning curves, reaching the highest success rates (e.g., $91.4\%$ in Alfworld; $100\%$ in Webshop) with stable gains. \textbf{Cumulative Return} is inconsistent; while competitive in Alfworld ($89.8\%$), it fails in Webshop ($\Delta \approx 0.8\%$), suggesting that absolute return targets can be brittle. \textbf{Single-episode Return} achieves relative improvements but lower final averages, indicating it lacks the cross-episode exploitation strength of the differential formulation.
\subsubsection{Population-Based Training}
\begin{wrapfigure}[25]{r}{0.45\textwidth} 
  \centering
  
  \includegraphics[width=0.42\textwidth]{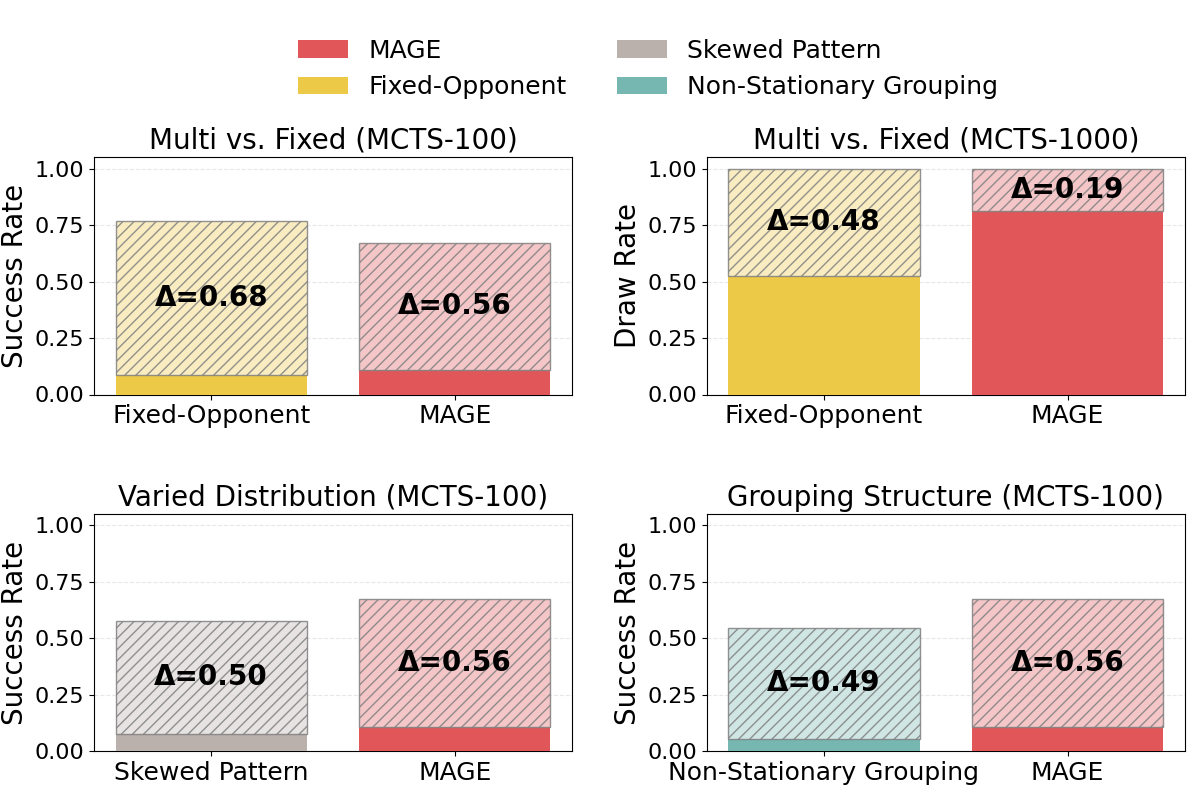}
  \caption{\textbf{Opponent diversity ablation results.} Using fixed-opponent, non-stationary grouping or skewed opponent weighting causes success rate to drop in Tic-Tac-Toe.}
\label{fig:opponent_ablation}
  \vspace{1.5em} 
\includegraphics[width=0.38\textwidth]{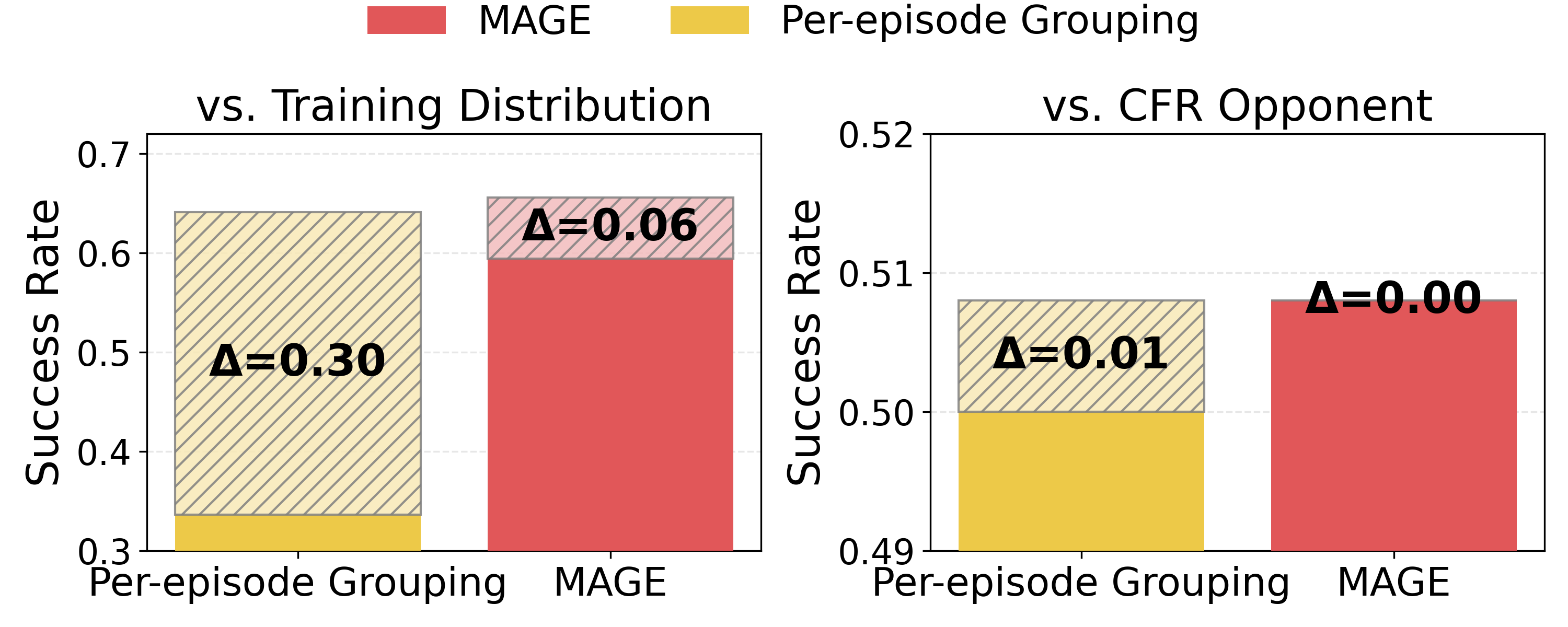}
\caption{\textbf{Advantage normalization ablation results.} Using per-episode grouping causes success rate to drop or fluctuate in Kuhn Poker.}
  \label{fig:adv_ablation}  
\end{wrapfigure}
We ablate the training distribution and sampling structure in Tic-Tac-Toe to assess their impact on meta-learning.\paragraph{Multi-Opponent vs. Fixed Opponent.}While the Fixed-Opponent baseline slightly outperforms MAGE against MCTS-100, this advantage is limited to task-specific memorization. When tested against MCTS-1000, MAGE demonstrates superior zero-shot generalization, achieving a $100\%$ success rate faster than the baseline. This confirms that MAGE’s multi-opponent training fosters more robust, scalable policies compared to the specialized, brittle strategies developed through single-opponent training.
\paragraph{Varied Distribution.}Comparing a \textit{Balanced Distribution} (50\% MCTS, 50\% patterns/random) to a \textit{Pattern-Skewed} version shows that MAGE (Balanced) outperforms the Skewed variant ($67.2\%$ vs. $57.8\%$). Exposure to diverse, patterned agents acts as a necessary curriculum for building robust opponent models.\paragraph{Grouping Structure.}We compare \textit{Stationary Grouping} (standard MAGE), where normalization groups share a single opponent archetype, against \textit{Non-Stationary Grouping} (mixed archetypes). Stationary grouping achieves $67.2\%$, while the non-stationary variant drops to $54.7\%$. By isolating archetype-specific variance, Stationary Grouping provides a cleaner credit assignment signal, enabling the agent to refine strategies effectively from interaction history.

\subsubsection{Agent-specific Advantage Normalization}
We evaluate two grouping strategies for GiGPO-style normalization:
\textbf{Cross-episode grouping (Global Anchor):} Aggregates all actions at state $s$ across the entire meta-episode into a single anchor group, $\mathcal{G}_{\text{global}}(s)$.
\textbf{Per-episode grouping (Local Anchor):} Normalizes actions at state $s$ only within each individual episode $n$, $\mathcal{G}_{n}(s)$.
The global anchor captures inter-episode dependencies for long-term improvement, while the local anchor preserves episode-specific context. 
MAGE (global anchor) outperforms the local variant in Kuhn Poker (Figure~\ref{fig:adv_ablation}), starting at $59.4\%$ and rapidly hitting the $65.6\%$ ceiling versus $33.6\%$. Global normalization stabilizes the baseline, linking early actions to exploitation. While both eventually reach the $\approx 50.8\%$ CFR theoretical limit, MAGE minimizes variance and ensures more consistent policy updates.

\section{Conclusion}
\label{sec:conclusion}
In this work, we introduced MAGE, a meta-reinforcement learning framework designed to equip Large Language Model agents with the capability for strategic exploration and exploitation in multi-agent environments. By shifting the paradigm from static execution to dynamic adaptation, MAGE enables agents to actively identify and capitalize on opponent vulnerabilities through multi-episode interactions. Our integration of population-based training with agent-specific advantage normalization effectively addresses the challenges of opponent diversity and reward instability, fostering robust ``learning-to-learn'' behaviors. Empirical results demonstrate that MAGE not only outperforms existing baselines in both exploration and exploitation tasks but also exhibits strong zero-shot generalization against unseen strategies. These findings highlight the necessity of internalizing meta-learning mechanisms within LLMs, paving the way for more autonomous agents capable of navigating the complexities of non-stationary, real-world interactions without relying on external scaffolding.
\section*{Impact Statement}
\label{sec:impact}



This paper presents work whose goal is to advance the field of meta-reinforcement learning for language agents. Our framework, MAGE, empowers large language model agents to perform strategic exploitation in multi-agent environments, which could be applied to areas such as adaptive educational tools and complex resource allocation scenarios. Our method may be adapted to other domains requiring rapid adaptation, like human-computer interaction. It is also essential to ensure the responsible deployment of our approach to avoid potential misuse.

\bibliography{reference}
\bibliographystyle{iclr2026_conference}

\newpage
\appendix
\onecolumn

\section{Reward Design Details}
\label{app:reward-design}

The total episode reward in our framework is composed of three main components: task reward, invalid action penalty, and length penalty.

\paragraph{Task reward.} The task reward reflects the primary objective of the episode. Successful outcomes (e.g., winning a game or completing a task correctly) are assigned a positive reward of 10, while failures (e.g., losing a game or incorrect completion) incur a negative reward of -10. Episodes that do not result in a clear success or failure receive a reward of 0.

\paragraph{Invalid action penalty.} To discourage the model from producing invalid or inadmissible actions, we apply a small invalid action penalty of 0.5.

\paragraph{Length penalty.} To control verbosity and encourage concise outputs, we impose a length-based penalty that gradually increases once the response length exceeds half of the maximum allowed length. 
\begin{equation}
r_{length} =
\begin{cases}
0, & L < \frac{1}{2} L_{\max},\\[6pt]
\dfrac{L - \frac{1}{2} L_{\max}}{L_{\max} - \frac{1}{2} L_{\max}}, & \frac{1}{2} L_{\max} \le L < L_{\max},\\[8pt]
1, & L \ge L_{\max},
\end{cases}
\end{equation}

Together, these reward components guide the model to produce correct, valid, and concise responses while maintaining a strong focus on task performance.

\section{Training and Evaluation Details}

\subsection{Shared Training Hyperparameters}
Unless otherwise specified, all experiments use the shared hyperparameters listed in Table~\ref{tab:shared_hyperparams}. 
We use an AdamW optimizer with a constant learning rate and employ GiGPO with mean-normalization for advantage stabilization.

\begin{table}[h]
\centering
\small
\begin{tabular}{lc}
\toprule
\textbf{Hyperparameter} & \textbf{Value} \\
\midrule
Actor Learning Rate & $1 \times 10^{-6}$ \\
PPO Mini-batch Size & 64 \\
PPO Micro-batch Size (per GPU) & 8 \\
Log-prob Micro-batch Size (per GPU) & 16 \\
Sampling Temperature & 0.7 \\
Top-$p$ / Top-$k$ & 0.8 / 20 \\
Invalid Action Penalty Coef. & 0.5 \\
Step-wise Discount ($\gamma_{\text{step}}$) & 0.95 \\
Trajectory Discount ($\gamma_{\text{traj}}$) & 0.6 \\
GiGPO Step Advantage Weight & 1.0 \\
GiGPO Normalization Mode & \texttt{mean\_norm} \\
Total Training Epochs & 150 \\
Evaluation Seed & 0 \\
\bottomrule
\end{tabular}
\caption{Shared training and rollout hyperparameters across all environments.}
\label{tab:shared_hyperparams}
\end{table}

\subsection{Environment-Specific Configurations}
Task-specific constraints and architectural parameters are detailed below. 
For all tasks, the maximum prompt and response lengths during testing remain consistent with training unless otherwise noted.

\paragraph{AlfWorld}
The environment uses \texttt{alfworld/AlfredTWEnv} with the standard \texttt{train} split for optimization. 
In-domain and out-of-domain evaluations utilize the \texttt{eval\_in\_distribution} and \texttt{eval\_out\_of\_distribution} sets, respectively. 
Parameters include 10 maximum turns, a maximum prompt length of 4096, a maximum response length of 1024, a maximum of 16384 batched tokens, and a reference log-probability micro-batch size of 16.

\paragraph{Sokoban}
Training is conducted over 300 epochs on $6 \times 6$ rooms with two boxes. 
Constraints include a maximum of 7 turns, a search depth of 100, a maximum of 21 solution steps, and 3 actions per turn. 
The configuration uses a maximum prompt length of 4096, a maximum response length of 4096, 32768 maximum batched tokens, and a length penalty coefficient of 1.0.

\paragraph{WebShop}
Optimization is performed over 150 epochs with a maximum of 12 turns, a maximum prompt length of 8192, a maximum response length of 1024, 32768 maximum batched tokens, and a reference log-probability micro-batch size of 32. 
For out-of-domain evaluation, the maximum prompt length is extended to 10240 tokens.

\paragraph{Tic-Tac-Toe}
This environment utilizes KL-divergence regularization (coefficient 0.1 with the \texttt{low\_var\_kl} variant) and a length penalty coefficient of 2.0. 
The setup includes 150 training epochs, a maximum of 8 turns, a $3 \times 3$ board, a maximum prompt length of 4096, a maximum response length of 3072, and 16384 maximum batched tokens.

\paragraph{Kuhn Poker}
Training consists of 150 epochs with a maximum of 6 turns. 
Parameters include a maximum prompt length of 4096, a maximum response length of 4096, 16384 maximum batched tokens, and a length penalty coefficient of 2.0.

\section{Prompt Templates}
\label{appendix:prompts}

This section provides the full prompt templates used for the decision-making (Play) and the reflection (Reflect) stages across all evaluated environments. Placeholders such as \texttt{\{init\_observation\}} and \texttt{\{current\_trajectory\}} are dynamically populated during the interaction.

\tcbset{
    promptstyle/.style={
        colback=gray!5,
        colframe=gray!50,
        fonttitle=\bfseries,
        breakable,          
        enhanced,
        left=3mm,
        right=3mm,
        top=3mm,
        bottom=3mm,
        fontupper=\small\ttfamily\raggedright 
    }
}

\subsection{ALFWorld Prompts}

\begin{tcolorbox}[promptstyle, title=ALFWorld Play Prompt]
You are an expert agent operating in the ALFRED Embodied Environment. \par\smallskip
\{init\_observation\}\{past\_trajectories\_reflections\}\{current\_trajectory\} \par\smallskip
Your admissible actions of the current situation are: \par
[\{admissible\_actions\}] \par\smallskip
Now it's your turn to take an action. \par\smallskip
- Your response should first by step-by-step reasoning about the current situation. \par
- Once you've finished your reasoning, you should choose an admissible action for current step and present it within <action> </action> tags.
\end{tcolorbox}

\begin{tcolorbox}[promptstyle, title=ALFWorld Reflect Prompt]
You are an expert agent operating in the ALFRED Embodied Environment. \par\smallskip
\{init\_observation\} \par\smallskip
You will be given the history of a past experience. Your job is to **reflect on the past sequence**, identify any **mistakes or inefficiencies**, and then devise a **concise, improved plan** starting from the original initial state. \par\smallskip
Below are the actions you took and the corresponding observations: \par
\{current\_trajectory\} \par
The task is NOT successfully completed. \par\smallskip
Now it's your turn to reflect on the past experience and come up with a new plan of action. \par\smallskip
- Your response should first be step-by-step reasoning about the strategy and path you took to attempt to complete the task. Identify where things went wrong or could be better. \par
- Then devise a concise, new plan of action that accounts for your mistake with reference to specific actions that you should have taken. \par
- Finally, end the response with your reflection and improved plan inside <remark> </remark> tags, to guide the next trial.
\end{tcolorbox}

\subsection{Sokoban Prompts}

\begin{tcolorbox}[promptstyle, title=Sokoban Play Prompt]
You are an expert agent operating in the Sokoban environment. \par\smallskip
\# Symbols and Their Meaning \par
- Walls (\#): These block movement. You can't move through or push anything into walls. \par
- Floor (\_): Open spaces where you can walk and move boxes. \par
- Targets (O): The spots where boxes need to go. \par
- Boxes (X): These are what you need to push onto the targets. \par
- Player (P): That's you! You'll move around the grid to push boxes. \par
- Box on Target ($\checkmark$): A box successfully placed on a target. \par
- Player on Target (S): You standing on a target. \par\smallskip
\# Goal \par
Your goal is to push all the boxes (X) onto the target spots (O). Once all boxes are on the targets, you win! \par\smallskip
\# Rules \par
Your admissible actions are ["up", "down", "left", "right"]. \par
You can only push one box at a time. You can't pull boxes, so plan ahead to avoid getting stuck. \par
You can't walk through or push boxes into walls (\#) or other boxes. \par
To avoid traps, do not push boxes into corners or against walls where they can't be moved again. \par\smallskip
\# Observations \par
The initial state of the game is: \par
\{init\_observation\}\{past\_trajectories\_reflections\}\{current\_trajectory\} \par
Now it's your turn to make moves (choose the next \{num\_actions\_per\_turn\} actions). \par\smallskip
- Your response first be step-by-step reasoning about the current situation — observe the positions of boxes and targets, plan a path to push a box toward a target, and avoid traps like corners or walls. \par
- Then choose \{num\_actions\_per\_turn\} admissible actions and present them within <action> </action> tags (separated by comma).
\end{tcolorbox}

\subsection{WebShop Prompts}

\begin{tcolorbox}[promptstyle, title=WebShop Play Prompt]
You are an expert autonomous agent operating in the WebShop e‑commerce environment. \par
Your task is to: \{task\_description\}.\{past\_trajectories\_reflections\}\{current\_trajectory\} \par\smallskip
Your admissible actions of the current situation are: \par
[ \{admissible\_actions\} ]. \par\smallskip
Now it's your turn to take one action for the current step. \par
Your response should first be step-by-step reasoning about the current situation, then think carefully which admissible action best advances the shopping goal. \par
Once you've finished your reasoning, you should choose an admissible action for current step and present it within <action> </action> tags.
\end{tcolorbox}

\subsection{Tic-Tac-Toe Prompts}

\begin{tcolorbox}[promptstyle, title=Tic-Tac-Toe Play Prompt]
You are an expert agent playing Tic-Tac-Toe on a \{board\_size\} by \{board\_size\} board. \par
The rows and columns are indexed from 1 to \{board\_size\}. \par\smallskip
\# Cell States \par
- Empty cells (.): cells that are not yet taken \par
- Player X (X): cells taken by player X \par
- Player O (O): cells taken by player O \par\smallskip
\# Game Rules \par
- A valid action is placing your mark on an empty cell (.). \par
- Each action is represented as a coordinate (row, col). \par
- At the INITIAL state, all of the following actions are valid: (1,1), (1,2), (1,3), (2,1), (2,2), (2,3), (3,1), (3,2), (3,3) \par
- A player WINS if they place THREE marks consecutively in a straight line (ROW, COLUMN, or DIAGONAL). \par\smallskip
\# Your Goal \par
Your goal is to win or prevent the opponent from winning. You play as \{player\_symbol\}, opponent plays as \{opponent\_symbol\}. \par\smallskip
\# Observation \par
\{init\_observation\}\{past\_trajectories\_reflections\}\{current\_trajectory\} \par\smallskip
IMPORTANT: The action history contains ONLY YOUR actions. Opponent acts automatically. \par
- Reason step-by-step about threats/opportunities. \par
- Choose ONE EMPTY cell (.) and put the index in "(row, col)" format within <action> </action> tags.
\end{tcolorbox}

\subsection{Kuhn Poker Prompts}

\begin{tcolorbox}[promptstyle, title=Kuhn Poker Play Prompt]
You are an expert Kuhn Poker agent. \par\smallskip
\# Game Rules \par
- Deck includes: King (K) > Queen (Q) > Jack (J). \par
- Both players place 1 chip blind ante. Each is dealt a private card. \par
- Players choose: PASS or BET (1 additional chip). \par
- If a player PASSES after a BET, the opponent wins. If both PASS or both BET, the higher card wins. \par\smallskip
\# Player Information \par
- You are playing as Player \{agent\_player\_id\}, opponent is Player \{opponent\_player\_id\}. \par\smallskip
\# Your Goal \par
Maximize your total chips over the long run. \par\smallskip
\# Observation \par
\{init\_observation\}\{past\_trajectories\_reflections\}\{current\_trajectory\} \par\smallskip
Now it's your turn: \par
- Reason step-by-step about the current situation. \par
- Choose ONE action: PASS or BET inside <action> </action> tags.
\end{tcolorbox}

\begin{table*}[h]
\centering
\footnotesize
\setlength{\tabcolsep}{4pt}
\renewcommand{\arraystretch}{1.15}
\begin{tabular}{p{0.48\linewidth} | p{0.48\linewidth}}
\toprule
\textbf{Reflection} & \textbf{Agent Response} \\
\midrule
\textit{Reflection summarizing errors and strategic guidance.} & \textit{Action generated by the agent after internalizing the reflection.} \\

The \textbf{key mistakes} in the past trial were:
1. Invalid moves (e.g., (-1,-1)) that disrupted the game state.
2. Poor initial strategy ...

\textbf{Improved Plan:}
Start with the center (2,2) to maximize control and opportunities for winning. ...
&
\texttt{</think>}

...
\textbf{The improved plan says} to start with the center (2,2) because that's the best move for maximizing control and opportunities. So first move should be (2,2). ...

\texttt{</think>}

\\
\bottomrule
\end{tabular}
\caption{Qualitative behavior analysis in Tic-Tac-Toe.}
\label{tab:behavior_analysis_ttt}
\end{table*}
\begin{table}[t]
\centering
\small
\renewcommand{\arraystretch}{1.15}
\setlength{\tabcolsep}{6pt}
\begin{tabular}{p{0.95\linewidth}}
\toprule
\textbf{Agent Response} \\
\midrule
\textit{The agent explicitly considers multiple plausible opponent strategies rather than assuming a fixed policy, and selects an action that is robust under strategic uncertainty.}\\
\texttt{</think>}\\
...
So, what's the possible scenario here?

Case 1: Player 0 has King. ...
Case 2: Player 0 has Jack. ...
So, in this case, \textbf{it's better to bet}. 

...

\textbf{Wait, but that's not necessarily the case.}
If Player 0 has a King, then they have a strong hand. If they bet, that's a strong move. But if Player 0 has a Jack, then \textbf{they might bet as well}. ...

\\
\texttt{</think>} \\
\bottomrule
\end{tabular}
\caption{Qualitative behavior analysis in Kuhn Poker illustrating multi-policy reasoning.}
\label{tab:behavior_multipolicy}
\end{table}
\section{Mechanistic Behavior Analysis}

To understand how MAGE shapes decision-making, we qualitatively analyze representative responses focusing on reflection-based learning and multi-opponent policy recognition.

\paragraph{Learn-to-Learn from Reflection.} 
To provide qualitative evidence that MAGE explicitly trains the agent to learn to learn rather than relying on emergent in-context learning, we analyze a representative interaction from the Tic-Tac-Toe environment.

As shown in Table~\ref{tab:behavior_analysis_ttt}, after an episode of errors (e.g., invalid actions, poor positioning), the agent consolidates these failures into a structured reflection. Unlike passive memory, MAGE optimizes the policy to exploit this feedback. In the subsequent episode, the agent demonstrates causally grounded correction: it reasons about prior mistakes, validates the action space, and selects the optimal center opening (2,2), translating reflection into immediate behavioral improvement.

\paragraph{Multi-opponent Policy Recognition and Generalization.}
Beyond learning from reflection, MAGE enables agents to adapt to environments involving strategic opponents with heterogeneous behaviors. This capability is particularly critical in multi-agent settings, where optimal actions depend not only on the current state but also on opponent policies.

In Kuhn Poker (Table~\ref{tab:behavior_multipolicy}), when facing a bet while holding a Queen, the agent does not assume a fixed adversarial strategy. Instead, it reasons over diverse opponent profiles—from aggressive to conservative—evaluating outcomes under multiple scenarios before acting. This behavior demonstrates internalized opponent modeling that is robust to policy variation rather than reliant on a single assumed equilibrium.
\begin{figure*}
    \centering
    \includegraphics[width=0.8\linewidth]{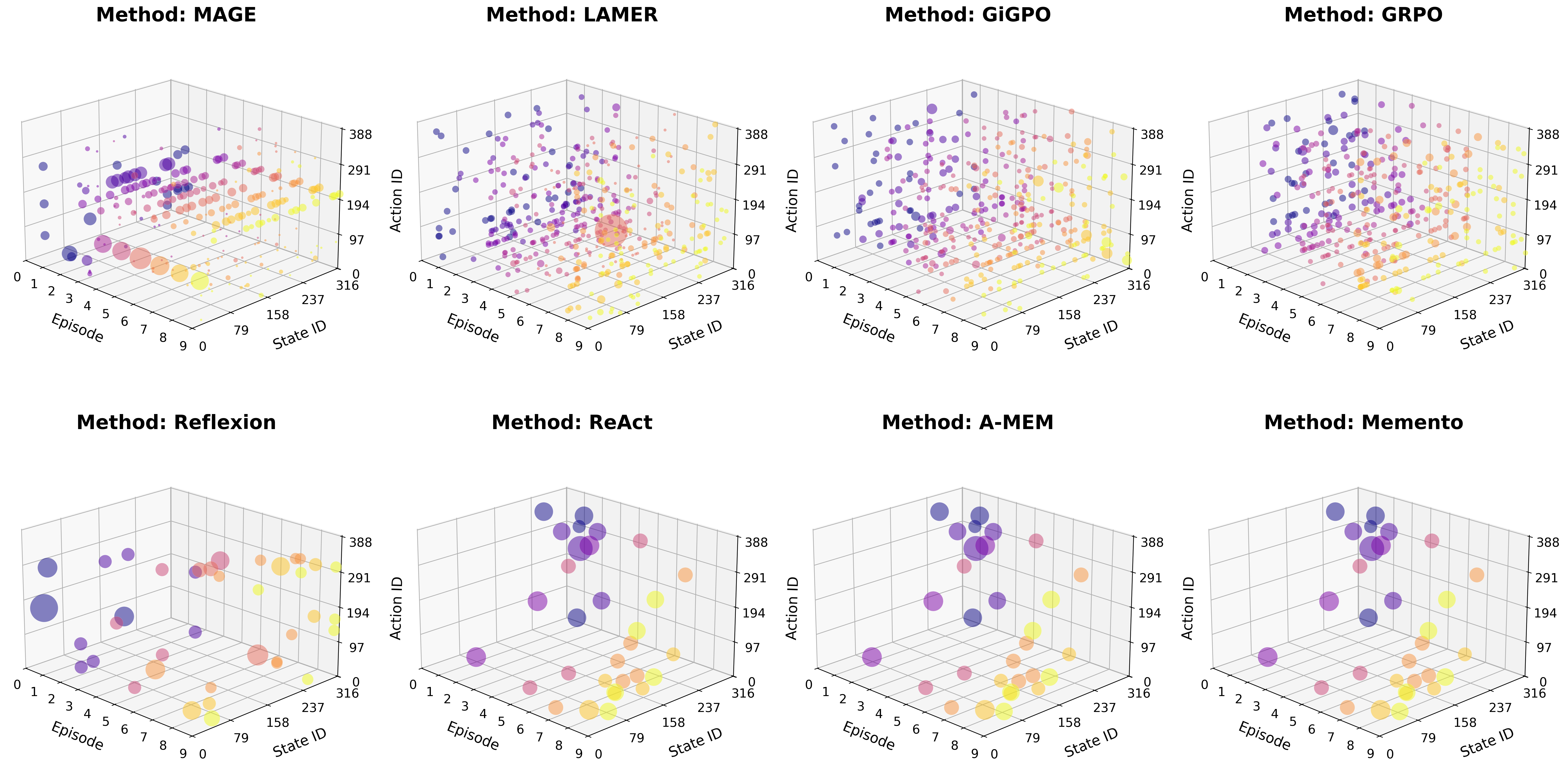}
    \caption{3D visualization of the state-action manifold across episodes for the WebShop task. The axes represent the Episode index ($X$), Shuffled State Index ($Y$), and Action Index ($Z$), with sphere size indicating the relative frequency of each $(s, a)$ pair.}
    \label{fig:webshop_scatter}
\end{figure*}
\section{Qualitative Analysis of Exploration and Policy Convergence}
To further investigate the internal dynamics of policy evolution, we visualize the state-action space across episodes in a 3D manifold ($X$: Episode, $Y$: State Index, $Z$: Action Index). The diameter of each sphere represents the frequency of a specific state-action pair $(s, a)$ within an episode.

\subsection{WebShop}
In the WebShop task, which requires complex multi-step reasoning and environment interaction, we observe distinct behavioral patterns (Figure~\ref{fig:webshop_scatter}).

MAGE exhibits a transition from broad exploration to structured exploitation. From $Episode \approx 4$, the state-action distribution becomes highly stabilized and regular. The spheres are distributed with remarkable homogeneity, suggesting that MAGE has identified a robust and repeatable trajectory. This visual regularity corresponds to its superior performance, where the success rate reaches $100.0\%$ at $Episode \ 4$ and maintains perfect execution thereafter. LAMER, GiGPO, GRPO methods show a persistently scattered distribution throughout the training process. The lack of concentrated "strategy tunnels" in the 3D space indicates that these agents fail to converge on an optimal path, resulting in suboptimal success rates.
The visualizations for Reflexion, ReAct, A-MEM, Memento methods are dominated by a few static, large circles with almost no surrounding exploration. This reflects a "frozen" policy that repeats fixed actions regardless of environmental feedback. Consequently, they suffer from premature stagnation, with success rates often flatlining near zero.

\begin{figure*}[!h]
    \centering
    \includegraphics[width=0.8\linewidth]{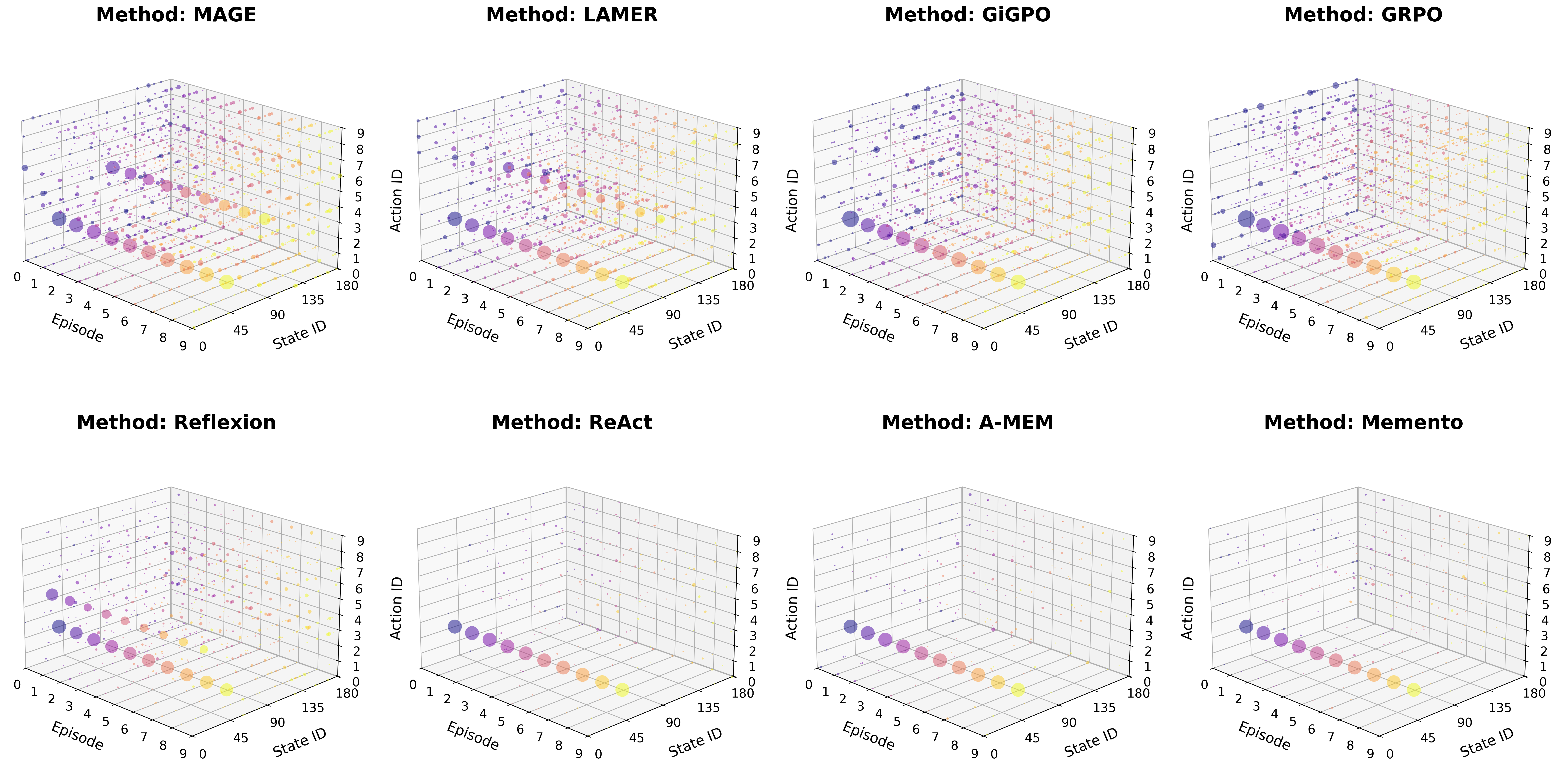}
    \caption{3D visualization of the state-action manifold across episodes for the Tic-Tac-Toe task. The axes represent the Episode index ($X$), Shuffled State Index ($Y$), and Action Index ($Z$), with sphere size indicating the relative frequency of each $(s, a)$ pair.}
    \label{fig:tictactoe_scatter}
\end{figure*}
\subsection{Tic-Tac-Toe}
The Tic-Tac-Toe task highlights the agent's ability to balance tactical focus with state-space coverage (Figure~\ref{fig:tictactoe_scatter}.

MAGE demonstrates a unique "dual-core" concentration, with two prominent state-action clusters representing key tactical responses. Crucially, these are accompanied by a wide and regular distribution of smaller spheres, indicating a healthy level of auxiliary exploration. This balanced profile allows MAGE to achieve the highest terminal success rate of $67.2\%$. LAMER shows a similar dual-cluster pattern but with a smaller secondary core compared to MAGE, leading to a slightly lower $60.2\%$. Reflexion also displays two clusters, yet its peripheral exploration (small dots) is significantly sparser, suggesting it lacks the necessary search breadth to handle diverse opponent moves. GiGPO, GRPO methods converge on only a single state-action cluster. While they show wide exploration, the dots are distributed irregularly, lacking the structured "policy lines" seen in MAGE. This implies that while they search the space, they fail to synthesize this experience into a coherent, multi-faceted strategy.
ReAct, A-MEM, Memento methods exhibit a singular large cluster with virtually no visible secondary points. Such extreme sparsity indicates a failure to explore alternative states, leading to near-zero success rates.

\subsection{Summary of Findings}
The 3D visualization confirms that MAGE's advantage stems from its ability to rapidly stabilize its core policy (observed as consistent "tunnels" along the episode axis) and maintain structured exploration (observed as the wide, regular distribution of smaller spheres), which prevents the rigid stagnation seen in baselines like ReAct or Reflexion.

\section{Limitations and Future Work} Although MAGE establishes a robust foundation for in-context adaptation, several promising directions remain for future exploration. While we focus on text-based environments, integrating multimodal feedback represents a natural evolution for cross-domain plasticity. Additionally, exploring dynamic, co-evolutionary training regimes where the opponent population evolves in response to the agent's progress could yield even more sophisticated strategic behaviors. Finally, while MAGE excels in discrete strategic tasks, evaluating its performance in open-ended, real-world environments with high-dimensional action spaces remains an important next step for verifying its broader applicability.
\section{Use of LLMs} We acknowledge the use of large language models (LLMs) to assist in the preparation of this manuscript. Specifically, LLMs are employed to improve the conciseness of the technical descriptions, ensure consistent LaTeX formatting across sections, and refine the grammatical flow of the analysis. 
\end{document}